\documentclass[runningheads]{llncs}
\usepackage{graphicx}
\usepackage[bottom]{footmisc}
\usepackage{amsmath,amssymb}
\usepackage{amssymb}
\usepackage{mathtools}
\usepackage{appendix}
\usepackage[square,numbers,sort&compress]{natbib}
\usepackage[section]{placeins}
\usepackage{longtable}
\usepackage{dirtytalk}
\usepackage{multirow}
\usepackage{subfig}
\usepackage{breqn}
\usepackage{xcolor}
\newcommand{\minimize}[2]{\ensuremath{\underset{\substack{{#1}}}{\textrm{minimize}}\;\;#2 }}
\def\bsq#1{
\lq{#1}\rq}

\usepackage{threeparttable}
\usepackage{color}

\begin{document}
\pagestyle{plain}
\title{DeConFuse : A Deep Convolutional Transform based Unsupervised Fusion Framework}
%
%
\author{Pooja Gupta\inst{1} \and Jyoti Maggu\inst{1} \and
Angshul Majumdar\inst{1,2} \and
Emilie Chouzenoux\inst{3} \and Giovanni Chierchia\inst{4}}
%
%
\institute{Indraprastha Institute of Information Technology, Delhi, India \and
TCS Research, Kolkata, India\\
 \and
Universit\'e Paris-Saclay, CentraleSup\'elec, Inria, CVN, Gif-sur-Yvette, France\\
\and 
LIGM, Universit\'e Gustave Eiffel, CNRS, ESIEE Paris, Noisy-le-Grand, France}
\maketitle              
\begin{abstract}
This work proposes an unsupervised fusion framework based on deep convolutional transform learning. The great learning ability of convolutional filters for data analysis is well acknowledged. The success of convolutive features owes to convolutional neural network (CNN). However, CNN cannot perform learning tasks in an unsupervised fashion. In a recent work, we show that such shortcoming can be addressed by adopting a convolutional transform learning (CTL) approach, where convolutional filters are learnt in an unsupervised fashion. The present paper aims at (i) proposing a deep version of CTL; (ii) proposing an unsupervised fusion formulation taking advantage of the proposed deep CTL representation; (iii) developing a mathematically sounded optimization strategy for performing the learning task. We apply the proposed technique, named DeConFuse, on the problem of stock forecasting and trading. Comparison with state-of-the-art methods (based on CNN and long short-term memory network) shows the superiority of our method for performing a reliable feature extraction. 
\end{abstract}

\section{Introduction}
In the last decade, Convolutional Neural Network (CNN) has enjoyed tremendous success in different types of data analysis. It was initially applied for images in computer vision tasks. The operations within the CNN were believed to mimic the human visual system. Although such a link between human vision and CNN may be present, it has been observed that deep CNNs are not exact models for human vision \cite{ref_1}. For instance, biologists consider that the human visual system would consist of 6 layers \cite{ref_2,ref_3} and not 20+ layers used in GoogleNet \cite{ref_4}.

Neural network models have also been used for analyzing time series data. Until recently, long short-term memory (LSTM) networks were the almost exclusively used neural network models for time series analysis as they were supposed to mimic memory and hence were deemed suitable for such tasks. However, LSTM are not able to model very long sequences, and their training is hardware intensive. Owing to these shortcomings, LSTMs are being replaced by CNNs. The reason for the great results of CNN methods for time series analysis (1D data processing in general) is not well understood. One possibility may lie in the universal function approximation capacity of deep neural networks \cite{ref_5,ref_6} rather than its biological semblance. The research in this area is primarily led by its success rather than its understanding. 

An important point to mention is that the performance of CNN is largely driven by the availability of very large labeled datasets. This probably explains their tremendous success in facial recognition tasks. Google's FaceNet \cite{ref_7} and Facebook's DeepFace \cite{ref_8} architectures are trained on 400 million facial images, a significant proportion of world’s population. These companies are easily equipped with gigantic labeled facial images data as these are \bsq{tagged} by their respective users. In the said problem, deep networks reach almost 100\% accuracy, even surpassing human capabilities. However, when it comes to tasks that require expert labeling, such as facial recognition from sketches (requiring forensic expertise) \cite{ref_8} or ischemic attack detection from EEG (requiring medical expertise) \cite{ref_9}, the accuracies become modest. Indeed, such tasks require expert labeling that is difficult to acquire, thus limiting the size of available labeled dataset. 

The same is believed by a number of machine learning researchers, including Hinton himself, are wary of supervised learning. In an interview with Axios,\footnotemark[1] Hinton mentioned his \bsq{deep suspicion} on backpropagation, the workhorse behind all supervised deep neural networks. He even added that \say{I don't think it's how the brain works,} and \say{We clearly don't need all the labeled data}. It seems that Hinton is hinting towards unsupervised learning frameworks. Unsupervised Learning technique does not require targets / labels to learn from data. This approach typically takes benefit from the fact that data is inherently very rich in its structure, unlike targets that are sparse in nature. Thus, it does not take into account the task to be performed while learning about the data, saving from the need of human expertise that is required in supervised learning. More on the topic of unsupervised versus supervised learning can be found in a blog by DeepMind.\footnotemark[2]
\footnotetext[1]{https://www.axios.com/artificial-intelligence-pioneer-says-we-need-to-start-over-1513305524-f619efbd-9db0-4947-a9b2-7a4c310a28fe.html}
\footnotetext[2]{https://deepmind.com/blog/article/unsupervised-learning} 

In this work, we would like to keep the best of both worlds, i.e. the success of convolutive models from CNN and the promises of unsupervised learning formulations. With this goal in mind, we developed convolutional transform learning (CTL) \cite{ref_10}. This is a representation learning technique that learns a set of convolutional filters from the data without label information. Instead of learning the filters (by backpropagating) from data labels, CTL learns them by minimizing a data fidelity loss, thus making the technique unsupervised. CTL has been shown to outperform several supervised and unsupervised learning schemes in the context of image classification. In the present work, we propose to extend the shallow CTL version to deeper layers, with the aim to generate a feature extraction strategy that is well suited for 1D time series analysis. This is the first major contribution of this work - deep convolutional transform learning.

In most applications, time series signals are multivariate, as they arise from multiple sources/sensors. For example, biomedical signals like ECG and EEG come from multiple leads; financial data from stocks are recorded with different inputs (open, close, low, high and net asset value), demand forecasting problems in smartgrids come with multiple types of data (power consumption, temperature, humidity, occupancy, etc.). In all such cases, the final goal is to perform prediction/classification task from such multivariate time series. We propose to address such problem as one of feature fusion. The information from each of the sources will be processed by the proposed deep CTL pipeline, and the generated deep features will be finally fused by an unsupervised fully connected layer. This is the second major contribution of this work - an unsupervised fusion framework with deep CTL. 

The resulting features can be used for different applicative tasks. In this paper, we will focus on the applicative problem of financial stock analysis. The ultimate goal may be either to forecast the stock price (regression problem) or to decide whether to buy or sell (classification problem). Depending on the considered task, we can pass the generated features into suitable machine learning tool, that may not be as data hungry as deep neural networks. Therefore, by adopting such a processing architecture, we expect to yield better results than traditional deep learning especially in cases where access to labeled data is limited.

\section{Literature Review}
\subsection{CNN for Time Series Analysis}
Let us briefly review and discuss CNN based methods for time series analysis. For a more detailed review, the interested reader can peruse \cite{ref_22}. We mainly focus on studies on stock forecasting as it will be our use case for experimental validation.

The traditional choice for processing time series with neural network is to adopt a recurrent neural network (RNN) architecture. Variants of RNN like long-short term memory (LSTM) \cite{LSTM_Hochreiter} and gated recurrent unit (GRU) \cite{GRU_Chung} have been proposed. 
However, due to the complexity of training such networks via backpropagation through time, they have been progressively replaced with 1D CNN \cite{ref_11}. For example, in \cite{ref_12}, a generic time series analysis framework was built based on LSTM, with assessed performance on the UCR time series classification datasets \cite{ref_14}. The later study from the same group \cite{ref_13}, based on 1D CNN, showed considerable improvement over the prior model on the same datasets. 

There are also several studies that convert 1D time series data into a matrix form so as to be able to use 2D CNNs~\cite{ref_15,ref_16,ref_17}. Each column of the matrix corresponds to a subset of the 1D series within a given time window and the resulting matrix is processed as an image. The 2D CNN model has been especially popular in stock forecasting. In \cite{ref_17}, the said techniques have been used on stock prices for forecasting. A slightly different input is used in \cite{ref_18}: instead of using the standard stock variables (open, close, high, low and NAV), it uses high frequency data for forecasting major points of inflection in the financial market. In another work \cite{ref_19}, a similar approach is used for modeling Exchange Traded Fund (ETF). It has been seen that the 2D CNN model performs the same as LSTM or the standard multi-layer perceptron \cite{ref_20,ref_21}. The apparent lack of performance improvement in the aforementioned studies may be due to an incorrect choice of CNN model, since an inherently 1D time series is modeled as an image. 

\subsection{Deep Learning and Fusion}

We now review existing works for processing multivariate data inputs, within the deep learning framework. Since the present work aims at being applied to stock price forecasting / trading, we will mostly focus our review on the multi-channel / multi-sensor fusion framework. Multimodal data and fusion for image processing, less related to our work, will be mentioned at the end of this subsection for the sake of completeness. 

Deep learning has been widely used recently for analyzing multi-channel / multi-sensor signals. In several of such studies, all the sensors are stacked one after the other to form a matrix and 2D CNN is used for analyzing these signals. For example, \cite{ref_23} uses this strategy for analyzing human activity recognition from multiple body sensors. It is important to distinguish such an approach from the aforementioned studies~ \cite{ref_17,ref_18,ref_19,ref_20,ref_21}. Here, the images are not formed from stacking windowed signals from the same signal one after the other, but by stacking signals from different sensors. The said study \cite{ref_23} does not account for any temporal modeling; this is rectified in \cite{ref_24}. In there, 2D CNN is used on a time series window; but the different windows are finally processed by GRU, thus explicitly incorporating time series modeling. 
There is however no explicit fusion framework in \cite{ref_23,ref_24}. The information from raw multivariate signals is simply fused to form matrices and treated by 2D convolutions. A true fusion framework was proposed in \cite{ref_25}. Each signal channel is processed by a deep 1D CNN and the output from the different signal processing pipelines are then fused by a fully connected layer. Thus, the fusion is happening at the feature level and not in the raw signal level as it was in \cite{ref_23,ref_24}. 

Another area that routinely uses deep learning based fusion is multi-modal data processing. This area is not as well defined as multi-channel data processing; nevertheless, we will briefly discuss some studies on this topic. In \cite{ref_26} a fusion scheme is shown for audio-visual analysis that uses a fusion scheme for deep belief network (DBN) and stacked autoencoder (SAE) for fusing audio and video channels. Each channel is processed separately and connected by a fully connected layer to produce fused features. These fused features are further processed for inference. We can also mention the work on video based action recognition addressed in \cite{ref_27}, which proposes a fusion scheme for incorporating temporal information (processed by CNN) and spatial information (also processed by CNN). 

There are several other such works on image analysis \cite{ref_28,ref_29,ref_30}. In \cite{ref_28}, a fusion scheme is proposed for processing color and depth information (via 3D and 2D convolutions respectively) with the objective of action recognition. In \cite{ref_29}, it was shown that by fusing hyperspectral data (high spatial resolution) with Lidar (depth information), better classification results can be achieved. In \cite{ref_30}, it was shown that by fusing deeply learnt features (from CNN) with handcrafted features via a fully connected layer, can improve analysis tasks. In this work, our interest lies in the first problem; that of inference from 1d / time-series multi-channel signals. To the best of our knowledge, all prior deep learning based studies on this topic are supervised. In keeping with the vision of Hinton and others, our goal is to develop an unsupervised fusion framework using deeply learn convolutive filters.     

\subsection{Convolutional Transform Learning}
Convolutional Transform Learning (CTL) has been introduced in our seminal paper \cite{ref_10}. Since it is a recent work, we present it in detail in the current paper, to make it self-content. CTL learns a set of filters $\left(t_m\right)_{1 \leq m \leq M}$ operated on observed samples $\left(s^{(k)}\right)_{1 \leq k \leq K}$ to generate a set of features $\big(x_m^{(k)}\big)_{1 \leq m \leq M,1 \leq k \leq K}$. Formally, the inherent learning model is expressed through convolution operations defined as
\begin{equation}
\label{eq1}
(\forall m \in \{ 1, \ldots,M\}\;, \forall k \in \{ 1, \ldots, K\})\qquad {t_m} * {s^{(k)}} = x_m^{(k)}.
\end{equation}

Following the original study on transform learning \cite{ref_37}, a sparsity penalty is imposed on the features for improving representation ability and limit overfitting issues. Moreover, in the same line as CNN models, the non-negativity constraint is imposed on the features.  
Training then consists of learning the convolutional filters and the representation coefficients from the data. This is expressed as the the following optimization problem
\begin{multline}
\label{eq2}
\minimize{{(t_m)_m},(x_m^{(k)})_{m,k}} \frac{1}{2}\sum\limits_{k = 1}^K {\sum_{m = 1}^M \Big( {\big\| {{t_m} * {s^{(k)}} - x_m^{(k)}} \big\|_2^2} }  + \psi( x_m^{(k)} ) \Big)
\\
+ \mu \sum_{m = 1}^M \left\| t_m \right\|_2^2 - \lambda \log \det \big( [ {{t_1}|\ldots|{t_M}} ] \big),
\end{multline}
where $\psi$ is a suitable penalization function. 
Note that the regularization term \say{$\mu \left\| \cdot \right\|_F^2 - \lambda \log \det$} ensures that the learnt filters are unique, something that is not guaranteed in CNN. Let us introduce the matrix notation
\begin{equation}
T*S-X
=
\begin{bmatrix}
t_1 * s^{(1)} - x_1^{(1)} & \dots & t_M * s^{(1)} - x_M^{(1)}\\
\vdots & \ddots&\vdots\\
t_1 * s^{(K)} - x_1^{(K)} & \dots & t_M * s^{(K)} - x_M^{(K)}\\
\end{bmatrix}
\end{equation}
where $T=\begin{bmatrix}t_1 & \dots & t_M\end{bmatrix}$, $S=\begin{bmatrix}s^{(1)} & \dots& s^{(K)}\end{bmatrix}^\top$, and $X=\begin{bmatrix} x_1^{(k)} & \dots & x_M^{(k)} \end{bmatrix}_{1\le k\le K}$. The cost function in Problem \eqref{eq2} can be compactly rewritten as\footnote{Note that $T$ is not necessarily a square matrix. By an abuse of notation, we define the \say{log-det} of a rectangular matrix as the sum of logarithms of its singular values.}
\begin{equation}
\label{eq:onelayer}
F(T,X) = \frac{1}{2}\left\| T*S - X \right\|_F^2 + \Psi(X) 
+ \mu \left\| T \right\|_F^2 - \lambda \log \det \left( T \right),
\end{equation}
where $\Psi$ applies the penalty term $\psi$ column-wise on $X$.

A local minimizer to \eqref{eq:onelayer} can be reached efficiently using the alternating proximal algorithm \cite{ref_34,ref_35,ref_36}, which alternates between proximal updates on variables $T$ and $X$. More precisely, set a Hilbert space $(\mathcal{H},\|\cdot\|)$, and define the proximity operator \cite{ref_21} at $\tilde x \in \mathcal{H}$ of a proper lower-semi-continuous convex function $\varphi : \mathcal{H} \to ] - \infty , + \infty ]$ as
\begin{equation}
\label{eq4}
\operatorname{prox}_{\varphi}(\tilde x) = \mathop {\arg \min }\limits_{x \in \mathcal{H}} \varphi (x) + \frac{1}{2}\left\| {x - \tilde x} \right\|^2.
\end{equation}
Then, the alternating proximal algorithm reads
\begin{equation}
\begin{array}{l}
{\rm{For\ }}{n} = 0,1,...\\
\left\lfloor \begin{array}{rl}
T^{[n + 1]} &= \operatorname{prox}_{\gamma_1 F(\cdot,X^{[n]})} \left(T^{[n]}\right)\\
X^{[n + 1]} &= \operatorname{prox}_{\gamma _2 F(T^{[n + 1]},\cdot)}\left( X^{[n]} \right)
\end{array} \right.
\end{array}
\end{equation}
with initializations $T^{[0]}$, $X^{[0]}$ and ${\gamma _1}, {\gamma _2}$ positive constants. For more details on the derivations and the convergence guarantees, the readers can refer to \cite{ref_10}.

\section{Fusion based on Deep Convolutional Transform Learning}
In this section, we discuss our proposed formulation. First, we extend the aforementioned CTL formulation to a deeper version. Next, we develop the fusion framework based on transform learning, leading to our DeConFuse\footnotemark[3] strategy.
\footnotetext[3]{Code available at: https://github.com/pooja290992/DeConFuse.git}

\subsection{Deep Convolutional Transform Learning}
\begin{figure}[!t]
\includegraphics[width=\textwidth]{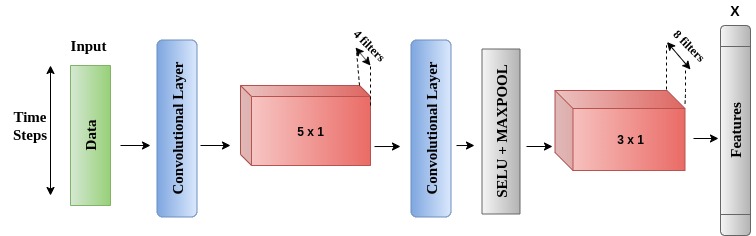}
\caption{Deep CTL architecture. The illustration is given for $L=2$ layers, with the first layer $T_1$ composed of $M_1=4$ filters of size $5\times1$, and the second layer composed of $M_2=8$ filters of size $3\times1$.}

\label{dctl}
\end{figure}

Deep CTL consists of stacking multiple convolutional layers on top of each other to generate the features, as shown in Figure \ref{dctl}. To learn all the variables in an end-to-end fashion, deep CTL relies on the key property that the solution $\widehat{X}$ to the CTL problem, assuming fixed filters $T$, can be reformulated as the simple application of an element-wise activation function, that is 
\begin{equation}\label{eq:relu}
\operatorname*{argmin}_X F(T,X) = \phi\big(T * S\big),
\end{equation}
with $\phi$ the proximity operator of $\Psi$ \cite{Combettes}. For example, if $\Psi$ is the indicator function of the positive orthant, then $\phi$ identifies with the famous rectified linear unit (ReLU) activation function. Many other examples are provided in \cite{Combettes}. Consequently, deep features can be computed by stacking many such layers
\begin{equation}\label{eq:deep_feats}
(\forall \ell\in\{1,\dots,L-1\})\qquad X_\ell = \phi_\ell(T_\ell *X_{\ell-1}),
\end{equation}
where $X_0 = S$ and $\phi_\ell$ a given activation function for layer $\ell$.

Putting all together, deep CTL amounts to
\begin{align}\label{eq:dctl}
\minimize{T_1,\dots,T_L,X}  F_{\rm conv}(T_1,\dots,T_L,X\,|\,S)
\end{align}
where
\begin{align}
F_{\rm conv}(T_1,\dots,T_L,X\,|\,S) &= \frac{1}{2} \| T_L*\phi_{L-1}(T_{L-1}*\dots \phi_1(T_1*S)) - X\|_F^2 \nonumber\\
&+ \Psi(X)  + \sum_{\ell=1}^{L}\big(\mu||T_\ell||^2_F - \lambda\log\det(T_\ell)\big).
\end{align}
This is a direct extension of the one-layer formulation in \eqref{eq:onelayer}. 

\subsection{Multi-Channel Fusion Framework}
\begin{figure}[!t]
\includegraphics[width=\textwidth]{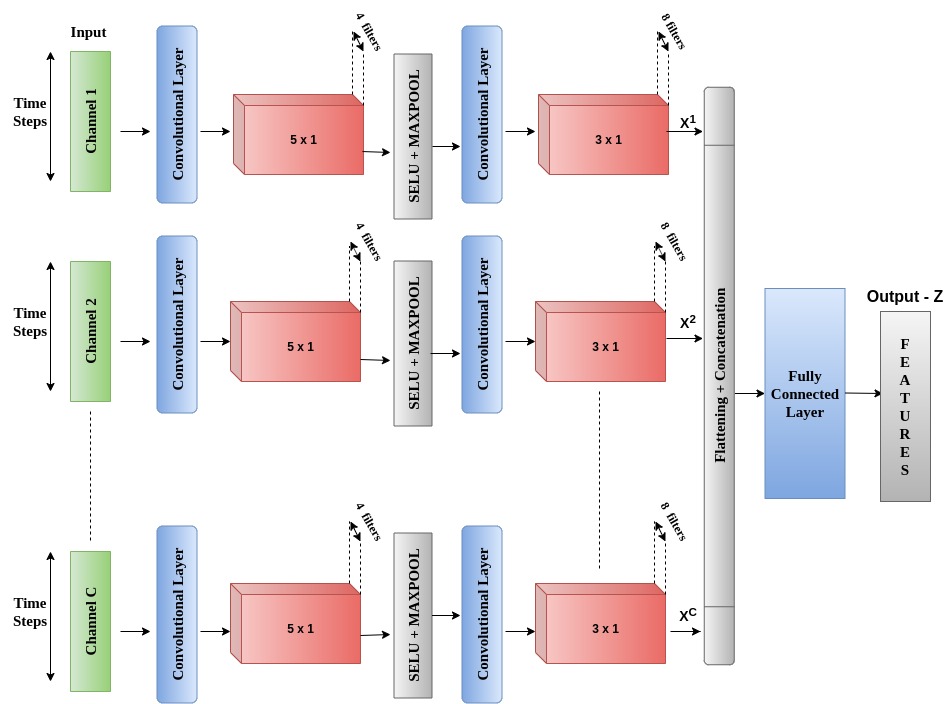}
\caption{DeConFuse architecture.}
\label{fdctl}
\end{figure}

We now propose a fusion framework to learn in an unsupervised fashion a suitable representation of multi-channel data that can then be utilised for a multitude of tasks. This framework takes the channels of input data samples to separate branches of convolutional layers, leading to multiple sets of channel-wise features. These decoupled features are then concatenated and passed to a fully-connected layer, which yields a unique set of coupled features. The complete architecture, called DeConFuse, is shown in Fig \ref{fdctl}. 

Since we have multi-channel data, for each channel $c \in \{1,\ldots,C\}$, we learn a different set of convolutional filters $T^{(c)}_1,\dots,T^{(c)}_L$ and features $X^{(c)}$. At the same time, we learn the (not convolutional) linear transform $\widetilde{T}=(\widetilde{T}_c)_{1\le c\le C}$ to fuse the channel-wise features $X = (X^{(c)})_{1\le c\le C}$, along with the corresponding fused features $Z$, which constitute the final output of the proposed DeConFuse model, as shown in Fig \ref{fdctl}. This leads to the joint optimization problem
\begin{equation}\label{eq:joint}
\minimize {T, X, \widetilde{T}, Z}  \underbrace{F_{\rm fusion}(\widetilde{T}, Z, X) +\sum_{c=1}^C F_{\rm conv}(T_1^{(c)},\dots,T_L^{(c)},X^{(c)}\,|\,S^{(c)})}_{J(T, X, \widetilde{T}, Z)}
\end{equation}
where
\begin{equation}
F_{\rm fusion}(\widetilde{T}, Z, X) = \frac{1}{2} \Big\|Z - \sum_{c=1}^C {\rm flat}(X^{(c)}) \widetilde{T}_c \Big\|^2_F + \iota_+(Z) + \sum_{c=1}^C\big(\mu\|\widetilde{T}_c\|^2_F - \lambda\log\det(\widetilde{T}_c)\big),
\end{equation}
where the operator \say{$\operatorname{flat}$} transforms $X^{(c)}$ into a matrix where each row contains the ``flattened'' features of a sample.

To summarize, our formulation aims to jointly train the channel-wise convolutional filters $T_\ell^{(c)}$ and the fusion coefficients $\widetilde{T}$ in an end-to-end fashion. We explicitly learn the features $X$ and $Z$ subject to non-negativity constraints so as to avoid trivial solutions and make our approach completely unsupervised. Moreover, the \say{log-det} regularization on both $T_\ell^{(c)}$ and $\widetilde{T}$  breaks symmetry and forces diversity in the learnt transforms, whereas the Frobenius regularization ensures that the transform coefficients are bounded.

\subsection{Optimization algorithm}\label{sec:optimization}
As for the solution of Problem \eqref{eq:joint}, we remark that all terms of the cost function are differentiable, except the indicator function of the non-negativity constraint. We can, therefore, find a local minimizer to \eqref{eq:joint} by employing the projected gradient descent, whose iterations read
\begin{equation}
\begin{array}{l}
{\rm{For\ }}{n} = 0,1,...\\
\;\left\lfloor \begin{array}{rl}
T^{[n + 1]} &= T^{[n]} - \gamma \nabla_T J(T^{[n]}, X^{[n]}, \widetilde{T}^{[n]}, Z^{[n]})\\
X^{[n + 1]} &= \mathcal{P}_+\big(X^{[n]} - \gamma\nabla_X J(T^{[n]}, X^{[n]}, \widetilde{T}^{[n]}, Z^{[n]})\big)\\
\widetilde{T}^{[n + 1]} &= \widetilde{T}^{[n]} - \gamma\nabla_{\widetilde{T}} J(T^{[n]}, X^{[n]}, \widetilde{T}^{[n]}, Z^{[n]})\\
Z^{[n + 1]} &= \mathcal{P}_+\big( Z^{[n]} - \gamma\nabla_Z J(T^{[n]}, X^{[n]}, \widetilde{T}^{[n]}, Z^{[n]})\big)\\
\end{array} \right.
\end{array}
\end{equation}
with initialization $T^{[0]}, X^{[0]}, \widetilde{T}^{[0]}, Z^{[0]}$,  $\gamma>0$, and $\mathcal{P}_+ = \max\{\cdot,0\}$. In practice, we make use of accelerated strategies \cite{ConvergenceAdam} within each step of this algorithm to speed up learning.

There are two notable advantages with the proposed optimization approach. Firstly, we rely on automatic differentiation \cite{AutoDifferentialNIPS} and stochastic gradient approximations to efficiently solve Problem \eqref{eq:joint}. Secondly, we are not limited to ReLU activation in \eqref{eq:deep_feats}, but rather we can use more advanced ones, such as SELU \cite{KlambauerSelfNorm}. This is beneficial for the performance, as shown by our numerical results.

\subsection{Computational Complexity of Proposed Framework - DeConFuse}
Table \ref{tab:complexity} summarizes the computational complexity of DeconFuse architecture, both for training and test phases. Specifically, it is reported the cost incurred for every input sample at each iteration of gradient descent in the training phase, and for the output computation in testing phase. The computational complexity of DeConFuse architecture is comparable to a regular CNN. The only addition is the log-det regularization, which requires to compute the truncated singular value decomposition of $T_\ell^{(c)}$ and $\widetilde{T}_c$. However, as the size of these matrices is determined by the filter size, the number of filters, and the number of output features per sample, the training complexity is not worse than that of a CNN.

\begin{table}
    \caption{\textbf{Time complexity in training and test phases (for one input sample)}}
    \label{tab:complexity}
    \centering
    \begin{tabular}{|c|l|c|l|}
    \hline
         \textbf{Phase} &
         \textbf{Steps} & 
         \textbf{Time Complexity}&
         \textbf{Dimension}
         \\
         & & & \textbf{Description}
         \\
         \hline
         \textbf{Training}
         &
         1. Convolution layers
         &
         $\mathcal{O}(P_\ell D_\ell M_\ell C)$ 
         &
         \\
         \textbf{phase} 
         & 
         2. Fully-connected (f.-c.) layer
         & 
         $\mathcal{O}(I^2 C^2)$ 
         &
         {\scriptsize $S^{(c)} \in \mathbb{R}^{K\times D}$}
         \\
         & 3. Frobenius norm on conv.\ layers & $\mathcal{O}\big(P_\ell M_\ell C\big)$
         & {\scriptsize $T_\ell^{(c)}\in\mathbb{R}^{P_\ell\times M_\ell}$}
         \\
         & 4. Frobenius norm on f.-c.\ layer 
         & $\mathcal{O}(I^2 C^2)$ 
         & {\scriptsize ${\rm flat}(X^{(c)})\in\mathbb{R}^{K\times I}$}
         \\
         & 5. log-det on conv.\ layers
         & $\mathcal{O}(P_\ell^2 M_\ell C)$ 
         & {\scriptsize $\widetilde{T_c} \in\mathbb{R}^{I \times O}$}
         \\
         & 6. log-det on f.-c.\ layer & $\mathcal{O}(I^3C^2)$ &
         \\
         \hline 
         \begin{tabular}[h]{c}
         \textbf{Testing} \\
         \textbf{phase} 
         \end{tabular}
         & 
         \begin{tabular}[h]{l}
         Step 1. + Step 2.
         \end{tabular}
         & 
         \begin{tabular}[h]{c}
         Step 1. + Step 2.
         \end{tabular}
         &
         \\ 
    \hline
    \end{tabular}
    \begin{tablenotes}
        \item {\scriptsize $D$ = input sample size -- $K$ = num.\ of samples -- $C$ = num.\ of channels -- $L$ = num.\ of layers}
        \item{\scriptsize $P_\ell$ = filter size at layer $\ell$ -- $M_\ell$ = num.\ of filters at layer $\ell$ -- $D_\ell$ = output sample size at layer $\ell$}
        \item {\scriptsize $I = D_L M_L$ is the num.\ of output features per sample and per channel at last convolution layer}
        \item {\scriptsize $O=\alpha IC$ (with $\alpha \in]0,1]$) is the num.\ of output features per sample at the fully-connected layer}
        %
        %
        %
    \end{tablenotes}
\end{table}

\section{Experimental Evaluation}
We carry out experiments on the real world problem of stock forecasting and trading. The problem of stock forecasting is a regression problem aiming at estimating the price of a stock at a future date (next day for our problem) given inputs till the current date. Stock trading is a classification problem, where the decision whether to buy or sell a stock has to be taken at each time. The two problems are related by the fact that simple logic dictates that if the price of a stock at a later date is expected to increase, the stock must be bought; and if the stock price is expected to go down, the stock must be sold. 

We will use the five raw inputs for both the tasks, namely open price, close price, high, low and net asset value (NAV). One could compute technical indicators based on the raw inputs \cite{ref_17} but, in keeping with the essence of true representation learning, we chose to stay with those raw values. Each of the five inputs is processed by a separate 1D processing pipeline. Each of the pipelines produces a flattened output (Fig. \ref{dctl}). The flattened outputs are then concatenated and fed into the Transform Learning layer acting as the fully connected layer (Fig. \ref{fdctl}) for fusion. While our processing pipeline ends here (being unsupervised), the benchmark techniques are supervised and have an output node. The node is binary (buy / sell) for classification and real valued for regression. More precisely, we will compare with two state-of-the-art time series analysis models, namely TimeNet \cite{ref_12} and ConvTimeNet \cite{ref_13}. In the former, the processing individual processing pipelines are based on LSTM and in the later they use 1D CNN.

We make use of a real dataset from the National Stock Exchange (NSE) of India. The dataset contains information of 150 symbols between 2014 and 2018; these stocks were chosen after filtering out stocks that had less than three years of data. The companies available in the dataset are from various sectors such as IT (e.g., TCS, INFY), automobile (e.g., HEROMOTOCO, TATAMOTORS), bank (e.g., HDFCBANK, ICICIBANK), coal and petroleum (e.g., OIL, ONGC), steel (e.g., JSWSTEEL, TATASTEEL),  construction (e.g., ABIRLANUVO, ACC), public sector units (e.g., POWERGRID, GAIL). The detailed architectures for each tested techniques, namely DeConFuse, ConvTimeNet and TimeNet are presented in the Table \ref{tab1}. For DeConFuse, TimeNet and ConvTimeNet, we have tuned the architectures to yield the best performance and have randomly initialized the weights for each stock's training.

\begin{table}[!htb]
\begin{threeparttable}
\caption{\textbf{Description of compared models}}\label{tab1}
\begin{tabular}{|c|p{6.9cm}|l|}
\hline
\textbf{Method} &  \textbf{Architecture Description} & \textbf{Other Parameters}\\
\hline
DeConFuse & 
\(
5 \times \begin{cases}
\textbf{layer1} : \textbf{1D Conv} (1,4,5,1,2)\tnote{1}\\
         \textbf{Maxpool} (2, 2)\tnote{2}\\
         \textbf{SELU}\\
\textbf{layer2} : \textbf{1D Conv} (5,8,3,1,1)\tnote{1}\\
\end{cases}\)
\newline
\newline
\textbf{layer3} : \textbf{Fully Connected}
& 
\begin{tabular}[h]{l}
Learning Rate = 0.001,\\
$\mu = 0.01, \epsilon = 0.0001$ \\
\textbf{Optimizer Used}: \textbf{Adam}\\
**with parameters**\\
$(\beta1,\beta2) = (0.9,0.999)$, \\
weight\_decay = 5e-5, \\
epsilon = 1e-8  
\end{tabular}
\\
\hline
ConvTimeNet & 
\( 
5 \times \begin{cases}
\textbf{layer1} : \textbf{1D Convolution} (1,32,9,1,4)\tnote{1}\\
         \textbf{Batch Normalization} + 
         \textbf{SELU}\\
\textbf{layer2} : \textbf{1D Convolution} (32,32,3,1,1)\tnote{1}\\
         \textbf{Batch Normalization} + 
         \textbf{SELU} + \textbf{SC}\tnote{3}\\

\textbf{layer3} : \textbf{1D Convolution} (32,64,9,1,4)\tnote{1}\\
         \textbf{Batch Normalization} + 
         \textbf{SELU}\\
\textbf{layer4} : \textbf{1D Convolution} (64,64,3,1,1)\tnote{1}\\
         \textbf{Batch Normalization} + 
         \textbf{SELU} + \textbf{SC}\tnote{3}\\
\textbf{layer3} : \textbf{Global Average Pooling}\\
\end{cases}\)
\newline
\newline
\textbf{layer4} : \textbf{Fully Connected}
\newline 
\textbf{For Trading, added} \textbf{layer5} : \textbf{Softmax}
&
\begin{tabular}[h]{l}
\textbf{For Forecasting:}\\
Learning Rate = 0.001,\\
\textbf{For Trading:}\\
Learning Rate = 0.0001,\\
\textbf{Optimizer Used}: \textbf{Adam}\\
**with parameters**\\
$(\beta1,\beta2) = (0.9,0.999)$, \\
weight\_decay = 1e-4, \\
epsilon = 1e-8  
\end{tabular}
\\
\hline
TimeNet & 
\(5 \times \begin{cases}
\textbf{layer1} : \textbf{LSTM unit} (1,12,2,True)\tnote{4}\\
\textbf{layer2} : \textbf{Global Average Pooling}\\
\end{cases}
\)
\newline
\newline
\textbf{layer3} : \textbf{Fully Connected}
\newline 
\textbf{For Trading, added}
\textbf{layer4} : \textbf{Softmax}
&
\begin{tabular}[h]{l}
\\
\textbf{For Forecasting:}\\
Learning Rate = 0.001,\\
\textbf{For Trading:}\\
Learning Rate = 0.0005,\\
\textbf{Optimizer Used}: \textbf{Adam}\\
**with parameters**\\
$(\beta1,\beta2) = (0.9,0.999)$, \\
weight\_decay = 5e-5, \\
epsilon = 1e-8  
\end{tabular}
\\
\hline
\end{tabular}
\begin{tablenotes}
\item[1] \small{(in_planes, out_planes, kernel\_size, stride, padding)}
\item[2] \small{(kernel\_size, stride)}
\item[3] \small{SC - Skip-Connection}
\item[4] \small{(input\_size,hidden\_size,\#layers,bidirectional)}
\end{tablenotes}
\end{threeparttable}
\end{table}


\subsection{Stock Forecasting – Regression}
 Let us start with the stock forecasting problem. We feed the generated unsupervised features from the proposed architecture into an external regressor, namely ridge regression. 
 Evaluation is carried out in terms of mean absolute error (MAE) between the predicted and actual stock prices for all 150 stocks. The stock forecasting results are shown in Table \ref{forecast} in appendix section \ref{a_forecast}. The MAE for individual stocks are presented for each of close price, open price, high price, low price and net asset value. 
 
 


From Table \ref{forecast}, it can be seen that the MAE values reached for the proposed DeConFuse solution for the four first prices (open, close, high, low) are exceptionally good for all of the 150 stocks. Regarding NAV prediction, the proposed method performs extremely well for 128 stocks. For the remaining 22 stocks, there are 13 
stocks, highlighted in red, for which DeConFuse does not give the lowest MAE but it is still very close to the best results given by the TimeNet approach. 

For a concise summary of the results, the average values over all stocks are shown in Table \ref{tab3}. 
For a concise summary of the results, the average values over all stocks are shown in Table \ref{tab3}. 
\begin{table}[!htb]
\caption{\textbf{Summary of Forecasting Results}}\label{tab3}
\centering
\begin{tabular}{|c|c|c|c|c|c|}
\hline
\textbf{Method} & \textbf{Close} & \textbf{Open} & \textbf{High} & \textbf{Low} & \textbf{NAV}\\
\hline
DeConFuse & \textbf{0.016} & \textbf{0.007} & \textbf{0.012} &
\textbf{0.013} & \textbf{0.410} \\
\hline
ConvTimeNet & 1.550 & 1.550 & 1.530 & 1.560 & 2.350 \\
\hline
TimeNet & 0.295 & 0.295 & 0.294 & 0.295 & 0.511 \\
\hline
\end{tabular}
\end{table}
\newline

\begin{figure}[!htb]
\begin{tabular}{l}
\subfloat[ABIRLANUVO and ALBK]{\includegraphics[width = 5.0in, height = 1.7in]{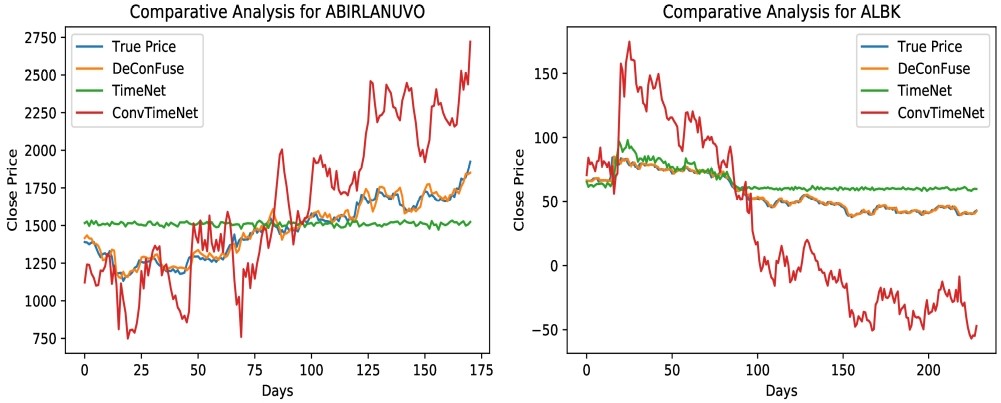}} \\
\subfloat[JPASSOCIAT and JSWENERGY]{\includegraphics[width = 5.3in, height = 2.6in]{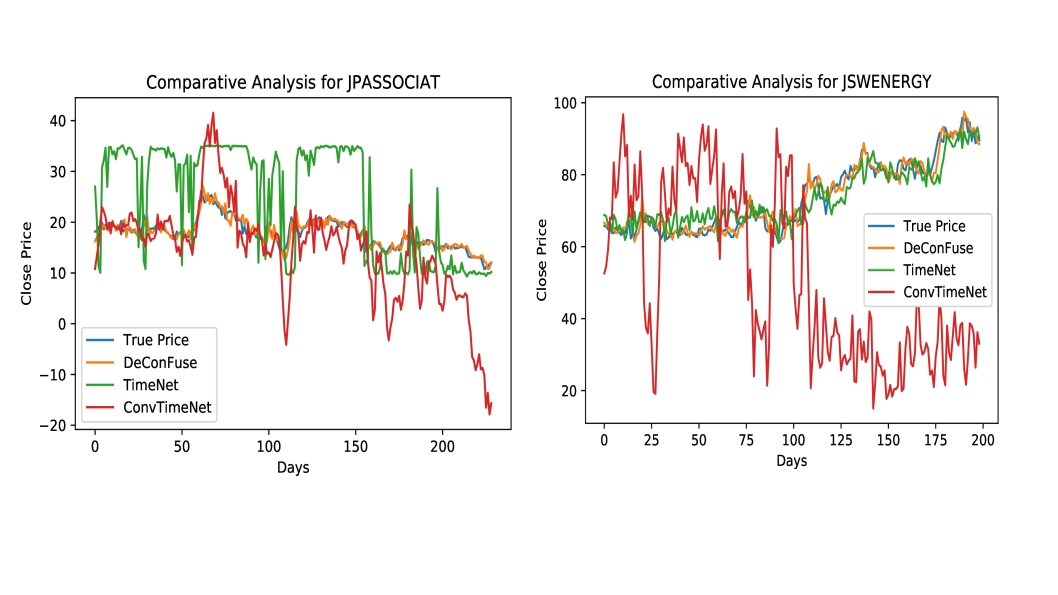}}
\end{tabular}
\caption{Stock Forecasting Performance}\label{close_plots}
\end{figure}

From the summary Table \ref{tab3}, it can be observed that our error is more than an order of magnitude better than the state-of-the-arts. The plots for one of the regressed prices (close price) for some examples of stocks in Fig. \ref{close_plots} show that the predicted close prices from DeConFuse are closer to the true close prices than benchmarks predictions.

\subsection{Stock Trading – Classification}
We now focus on the stock trading task. In this case, the generated unsupervised features from DeConFuse are inputs to an external classifier based on Random Decision Forest (RDF) with $5$ decision tree classifiers and depth $3$. Even though we used this architecture, we found that the results from RDF are robust to changes in architecture. This is a well known phenomenon about RDFs \cite{RDF_Criminisi}. We evaluate the results in terms of precision, recall, F1 score, and area under the ROC curve (AUC). From the financial viewpoint, we also calculate annualized returns (AR) using the predicted trading signals / labels as well as using true trading signals / labels named as Predicted AR and True AR respectively. The starting capital used for calculating AR values for every stock is Rs. 1,00,000 and the transaction charges are Rs 10. The stock trading results are shown in the Table \ref{trading} in   appendix section \ref{a_trading}.  

\begin{figure}[!htb]
\begin{tabular}{l}
\subfloat[RELIANCE and ONGC]{\includegraphics[width = 5.0in, height = 1.7in]{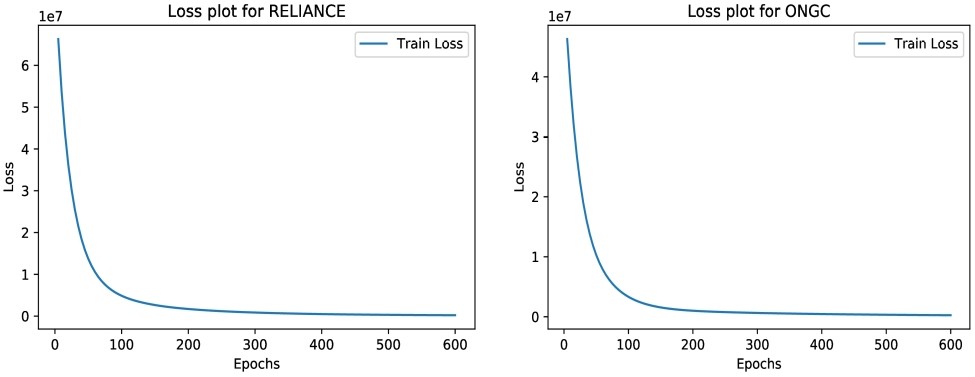}} \\
\subfloat[HINDUNILVR and ICICIBANK]{\includegraphics[width = 5.0in, height = 1.7in]{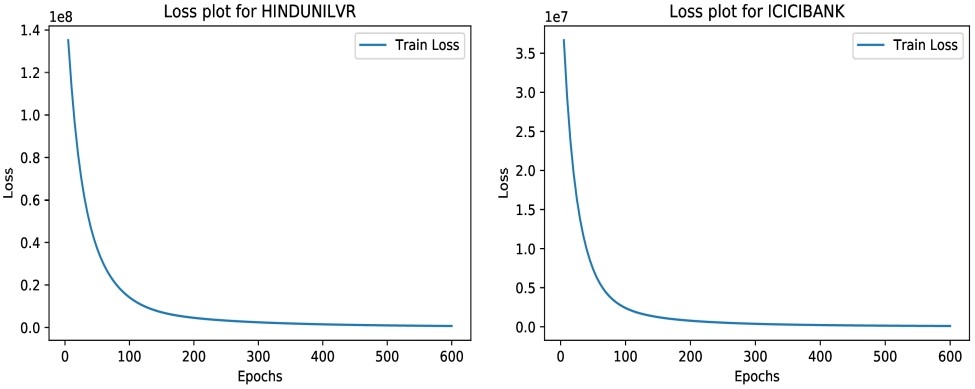}}
\end{tabular}
\caption{Empirical Convergence Plots}\label{loss_plots}
\end{figure}

\noindent
Certain results from Table \ref{trading} are highlighted in bold or red. The first set of results, marked in bold, are the ones where one of the techniques for each metric gives the best performance for each stock. The proposed solution DeConFuse gives the best results for 89 stocks for precision score, 85 stocks for recall score, 125 stocks for F1 score, 91 stocks for AUC measure, and 56 stocks in case of the AR metric. 
The other set marked in red highlights the cases where DeConfuse has not performed the best but performs nearly equal (here, a difference of maximum 0.05 in the metric is considered) to the best performance given by one of the benchmarks i.e. DeConFuse gives the next best performance. 
We noticed that there are 24 stocks for which DeConFuse gives the next best precision metric value. Likewise, 18 stocks in case of recall, 22 stocks for F1 score, 26 stocks for AUC values, and 1 stock in case of AR. Overall, DeConfuse reaches very satisfying performance over the benchmark techniques. This is also corroborated from the summary of trading results in Table \ref{tab5}.
\begin{table}[!htb]
\caption{\textbf{Summary of Trading Results}}\label{tab5}
\centering
\begin{tabular}{|c|c|c|c|c|c|}
\hline
\textbf{Method} & \textbf{Precision} & \textbf{Recall} & 
\begin{tabular}[t]{c}
\textbf{F1}\\
\textbf{Score}
\end{tabular}
& \textbf{AUC} & \begin{tabular}[t]{c}
\textbf{MAE}\\
\textbf{AR}
\end{tabular}
\\
\hline
DeConFuse & \textbf{0.520} & \textbf{0.810} & \textbf{0.628} &
\textbf{0.543} & \textbf{17.350} \\
\hline
ConvTimeNet & 0.510 & 0.457 & 0.413 & 0.524 & 19.410 \\
\hline
TimeNet & 0.470 & 0.648 & 0.490 & 0.513 & 18.760 \\
\hline
\end{tabular}
\end{table}
\\\\
We also display empirical convergence plots for few stocks, namely RELIANCE, ONGC, HINDUNILVR and ICICIBANK, in Fig.\ref{loss_plots}. We can see that the training loss decreases to a point of stability for each example. 

\section{Conclusion}
In this work, we propose DeConFuse, a deep fusion end-to-end framework for the processing of 1D multi-channel data. Unlike other deep learning models, our framework is unsupervised. It is based on a novel deep version of our recently proposed convolutional transform learning model. We have applied the proposed model for stock forecasting / trading leading to very good performance. The framework is generic enough to handle other multi-channel fusion problems as well. 

The advantage of our framework is its ability to learn in an unsupervised fashion. For example, consider the problem we address. For traditional deep learning based models, we need to retrain to deep networks for regression and classification. But we can reuse our features for both the tasks, without the requirement of re-training, for specific tasks. This has advantages in other areas as well. For example, one can either do ischemia detection, i.e. detect whether one is having a stroke at the current time instant (from EEG); or one can do ischemia prediction, i.e. forecast if a stroke is going to happen. In standard deep learning, two networks need to be retrained and tuned to tackle these two problems. With our proposed method, there is no need for this double effort. 

In the future, we would work on extending the framework for supervised / semi-supervised formulations. We believe that the semi-supervised formulation will be of immense practical importance. We would also like to extend it to 2D convolutions in order to handle image data. 

\section{Declarations}
\subsection{List of abbreviations}
\begin{itemize}
    \item TL : Transform Learning
    \item CTL : Convolutional Transform Learning 
    \item CNN : Convolutional Neural Network
    \item LSTM : Long Short Term Memory 
    \item GRU : Gated Recurrent Unit
    \item ReLU : Rectified Linear Unit
    \item SELU : Scaled Exponential Linear Units
    \item NSE : National Stock Exchange
    \item AUC : Area Under Curve
    \item ROC : Receiver Operating Characteristics 
    \item NAV : Net Asset Value
    \item RDF : Random Decision Forest 
    \item EEG : Electroencephalogram
    \item ECG : Electrocardiogram
    \item AR : Annualized Returns 
    \item MAE : Mean Absolute Error 
\end{itemize}

\subsection{Ethics approval and consent to participate}
Not Applicable.

\subsection{Consent for publication}
Not Applicable.

\subsection{Availability of data and materials}
The dataset used is a real dataset of the Indian National Stock Exchange (NSE) of past four years and is publicly available. We have shared the data with our implementation available at: https://github.com/pooja290992/DeConFuse.git.

\subsection{Competing interests}
The authors declare that they have no competing interests.

\subsection{Funding}
This work was supported by the CNRS-CEFIPRA project under grant NextGenBP PRC2017.

\subsection{Authors' contributions}
\begin{itemize}
    \item Ms. Pooja Gupta has introduced the CTL within the fusion framework and performed all the numerical experiments.
    \item Ms. Jyoti Maggu originally formulated the transform learning model and the deep version for it.
    \item Dr. Angshul Majumdar has helped with the model formulation and the assessment of the experimental part. 
    \item Dr. Emilie Chouzenoux and Dr. Giovanni Chierchia have contributed in the formulation of the model and the optimization algorithms. 
    \item All the authors have contributed to the writing and proofreading of the paper.
\end{itemize}

\newpage
\appendix
\vspace*{2.5in}
\section*{\qquad\qquad\qquad\qquad\Huge
{\textbf{Appendix}}}
\clearpage
\section{Detailed Stock Forecasting Results}
\label{a_forecast}
\begin{longtable}[!htb]{|c|c|c|c|c|c|c|}
\caption{Stock-wise Forecasting Results}\label{forecast}\\
\hline
\textbf{Stock Name} & \textbf{Method} & \begin{tabular}[t]{c}\textbf{ MAE } \\
\textbf{ Close } 
\end{tabular}
& \begin{tabular}[t]{c}\textbf{ MAE } \\
\textbf{ Open } 
\end{tabular} 
& 
\begin{tabular}[t]{c}\textbf{ MAE } \\
\textbf{ High } 
\end{tabular} 
& 
\begin{tabular}[t]{c}\textbf{ MAE } \\
\textbf{ Low } 
\end{tabular} 
& 
\begin{tabular}[t]{c}\textbf{ MAE } \\
\textbf{ NAV } 
\end{tabular} 
\\
\hline
\textbf{ABIRLANUVO} & DeConFuse & \textbf{0.021} & \textbf{0.015} & \textbf{0.019} & \textbf{0.017} & \textbf{0.416}\\
 & ConvTimeNet & 0.204 & 0.212 & 0.219 & 0.195 & 1.804\\
 & TimeNet & 0.112 & 0.111 & 0.111 & 0.112 & 0.467\\
 \hline
\textbf{ACC} & DeConFuse & \textbf{0.012} & \textbf{0.016} & \textbf{0.014} & \textbf{0.017} & 0.580\\
 & ConvTimeNet & 0.158 & 0.161 & 0.159 & 0.158 & 0.765\\
 & TimeNet & 0.116 & 0.116 & 0.115 & 0.118 & \textbf{0.388}\\
\end{longtable}

\section{Detailed Stock Trading Results}
\label{a_trading}
\begin{longtable}[!htb]{|c|c|c|c|c|c|c|c|}
\caption{\textbf{Stock-wise Trading Results}}\label{trading}\\
\hline
\multirow{2}{*}{\textbf{Stock Name}} &
\multirow{2}{*}{\textbf{Method}} &
  \multicolumn{4}{c|}{
  \begin{tabular}[t]{c}
  \textbf{ Computational }\\
  \textbf{ Model Performance }
  \end{tabular}
  } 
  &
  \multicolumn{2}{c|}
  {
  \begin{tabular}[t]{c}
  \textbf{ Financial }\\
  \textbf{ Evaluation }
  \end{tabular}
  } 
  \\
  \cline{3-8}
& & \textbf{Precision} & \textbf{Recall} & 
\begin{tabular}[t]{c}
\textbf{F1} \\
\textbf{Score} 
\end{tabular}
& \textbf{AUC} 
& 
\begin{tabular}[t]{c}
\textbf{True}\\
\textbf{AR} 
\end{tabular}
& 
\begin{tabular}[t]{c}
\textbf{Predicted}\\
\textbf{AR}
\end{tabular}
\\
\hline
\textbf{ABIRLANUVO} & DeConFuse & \textbf{0.553} & 0.886 &\textbf{ 0.681} & \textbf{0.558} & \textbf{41.950}
& \textbf{15.600}\\
 & ConvTimeNet & 0.515 & 0.966 & 0.672 & 0.541 &  & 3.090\\
 & TimeNet & 0.512 & \textbf{0.989} & 0.674 & 0.478 &  & 8.340\\
 \hline
\textbf{ACC} & DeConFuse & \textbf{0.449} & \textbf{0.761} & \textbf{0.565} & \textbf{0.600} & \textbf{-7.090} & -1.070\\
 & ConvTimeNet & \textbf{0.449} & 0.337 & 0.385 & 0.529 & & -4.100\\
\end{longtable}
\end{document}